\documentclass[11pt,a4paper]{article}
\usepackage[hyperref]{emnlp2020}
\usepackage{times}
\usepackage{latexsym}

\usepackage{microtype}
\usepackage{todonotes}
\usepackage{booktabs}
\usepackage{hyperref}
\usepackage{relsize}
\usepackage{tikz}
\usepackage{tabu} 
\usepackage{subcaption} 
\usepackage{mathtools}
\usepackage{amsmath}
\usepackage{xcolor,colortbl} 
\usepackage{float} 
\usepackage[export]{adjustbox} 

\usepackage{soul}

\aclfinalcopy 
\def\aclpaperid{1938} 

\newcommand{\bert}{\textsc{bert}}
\newcommand{\bertbase}{\textsc{bert-base}}
\newcommand{\bertonly}{\textsc{bert-only}}
\newcommand{\bertlarge}{\textsc{bert-large}}
\newcommand{\roberta}{\textsc{roberta}}
\newcommand{\robertabase}{\textsc{roberta-base}}
\newcommand{\robertalarge}{\textsc{roberta-large}}
\newcommand{\elmo}{\textsc{elmo}}
\newcommand{\glove}{\textsc{glove}}
\newcommand{\fasttext}{\textsc{fasttext}}
\newcommand{\drseval}{\textbf{\textsc{drs-jury}}}

\newcommand{\mytilde}{\raise.17ex\hbox{$\scriptstyle\mathtt{\sim}$}}

\newcommand{\gentag}[3][\normalsize]{\tikz[baseline=0ex]\node[
	draw=gray,
	fill=#3!20,
	inner sep=2.5pt, 
	anchor=text, 
	rectangle,
	rounded corners=.6mm]
{#1\smtg{#2}};} 
\newcommand{\smtg}[1]{\ensuremath{\texttt{#1}}\index{#1}} 

\newcommand{\semtag}[1]{\gentag{#1}{blue}}
\newcommand{\postag}[1]{\gentag{#1}{red}}
\newcommand{\lemtag}[1]{\gentag{#1}{yellow}}

\newcommand{\deptag}[1]{\gentag{#1}{green}}
\newcommand{\ccgtag}[1]{\gentag{#1}{orange}}

\newcommand{\clause}[1]{\texttt{#1}}
\newcommand{\disvar}[1]{\textbf{#1}}
\newcommand{\boxvar}[1]{\textbf{#1}}
\newcommand{\operator}[1]{#1}
\newcommand{\concept}[1]{#1}
\newcommand{\role}[1]{#1}

\newcommand{\vect}[1]{\mathbf{#1}}
\newcommand{\matr}[1]{\mathbf{#1}}

\definecolor{green}{rgb}{0.49, 0.99, 0.0}
\definecolor{red}{rgb}{1.0, 0.0, 0.0}
\definecolor{orange}{rgb}{1.0, 0.5, 0.0}
\newcommand{\cellgreen}{\cellcolor{green}}
\newcommand{\cellred}{\cellcolor{red}}
\newcommand{\cellorange}{\cellcolor{orange}}

\newcommand{\inlineheader}[1]{%
\vspace{0.06cm}
\noindent\textbf{#1}\quad
}
\newcommand{\inlineheadernoquad}[1]{%
\vspace{0.06cm}
\noindent\textbf{#1}
}

\title{Character-level Representations Improve DRS-based\\ Semantic Parsing Even in the Age of BERT}

\author{Rik van Noord \\
  CLCG \\
  University of Groningen \\
  The Netherlands \\
  \normalsize{\texttt{rikvannoord@gmail.com}} \\\And 

  Antonio Toral \\
  CLCG \\
  University of Groningen \\
  The Netherlands \\
  \normalsize{\texttt{a.toral.ruiz@rug.nl}} \\ \\\And
  
  Johan Bos \\
  CLCG \\
  University of Groningen \\
  The Netherlands \\
  \normalsize{\texttt{johan.bos@rug.nl}} \\}

\date{}

\begin{document}
\maketitle
\begin{abstract}
We combine character-level and contextual language model representations to improve performance on Discourse Representation Structure parsing. Character representations can easily be added in a sequence-to-sequence model in either one encoder or as a fully separate encoder, with improvements that are robust to different language models, languages and data sets. For English, these improvements are larger than adding individual sources of linguistic information or adding non-contextual embeddings. A new method of analysis based on semantic tags demonstrates that the character-level representations improve performance across a subset of selected semantic phenomena.
\end{abstract}

\section{Introduction}
\label{sec:intro}

Character-level models have obtained impressive performance on a number of NLP tasks, ranging from the classic POS-tagging \citep{santos2014learning} to complex tasks such as Discourse Representation Structure (DRS) parsing \citep{drstacl:18}. However, this was before the large pretrained language models \citep{Peters:2018,bert:19} took over the field, with the consequence that for most NLP tasks, state-of-the-art performance is now obtained by fine-tuning on one of these models (e.g., \citealp{conneau-etal-2020-unsupervised}).

Does this mean that, despite a long tradition of being used in language-related tasks (see Section~\ref{sec:backgroundchar}), character-level representations are no longer useful? We try to answer this question by looking at semantic parsing, specifically DRS parsing \citep{eacl:pmb, pmb-LREC:18}. We aim to answer the following research questions:

\begin{enumerate}

\setlength\itemsep{-4pt}

   \item Do pretrained language models (LMs) outperform character-level models for DRS parsing?

    \item Can character and LM representations be combined to improve performance, and if so, what is the best method of combining them? 
    
    \item How do these improvements compare to adding linguistic features?
    
    \item Are the improvements robust across different pretrained language models, languages, and data sets?
    
    \item On what type of sentences do character-level representations specifically help?
\end{enumerate}

\inlineheadernoquad{Why semantic parsing?} Semantic parsing is the task of automatically mapping natural language utterances to interpretable meaning representations. The produced meaning representations can then potentially be used to improve downstream NLP applications (e.g., \citealp{issa-etal-2018-abstract,song-etal-2019-semantic, mihaylov-frank-2019-discourse}), though the introduction of large pretrained language models has shown that explicit formal meaning representations might not be a necessary component to achieve high accuracy. However, it is now known that these models lack reasoning capabilities, often simply exploiting statistical artifacts in the data sets, instead of actually \textit{understanding} language \citep{niven-kao-2019-probing, mccoy-etal-2019-right}. Moreover, \citet{ettinger2020bert} found that the popular \bert{} model \citep{bert:19} completely failed to acquire a general understanding of negation. Related, \citet{bender-koller-2020-climbing} contend that meaning cannot be learned from form alone, and argue for approaches that focus on grounding the language (communication) in the real world. We believe formal meaning representations therefore have an important role to play in future semantic applications, as semantic parsers produce an explicit model of a real-world interpretation.

\begin{figure*}[!t]
\captionsetup{singlelinecheck=false,justification=justified}
\begin{minipage}[t]{.25\linewidth}
\vspace{0pt}
\setlength{\tabcolsep}{2pt}
\centering
\footnotesize
\begin{tabular}{ll}
\toprule
\multicolumn{2}{l}{\small \textbf{Sent:} I haven't been to Boston since 2013.} \\
\midrule
\clause{\boxvar{b1} \operator{NEGATION} \boxvar{b2}}      &\clause{\boxvar{b3} \operator{REF} \disvar{x1}}    \\
\clause{\boxvar{b1} \operator{REF} \disvar{t1}}    &\clause{\boxvar{b3} \role{Name} \disvar{x1} "boston"}    \\
\clause{\boxvar{b1} \operator{TPR} \disvar{t1} "now"}  &   \clause{\boxvar{b3} \operator{PRESUPPOSITION} \boxvar{b2}}    \\
\clause{\boxvar{b1} \concept{time} "n.08" \disvar{t1}}  &  \clause{\boxvar{b3} \concept{city} "n.01" \disvar{x1}}    \\
\clause{\boxvar{b2} \operator{REF} \disvar{e1}}    &\clause{\boxvar{b2} \role{Start} \disvar{e1} \disvar{t2} }    \\
\clause{\boxvar{b2} \role{Theme} \disvar{e1} "speaker"}  &   \clause{\boxvar{b2} \operator{REF} \disvar{t2}}    \\
\clause{\boxvar{b2} \role{Time} \disvar{e1} \disvar{t1}}  &  \clause{\boxvar{b2} \concept{time "n.08"} \disvar{t2}}    \\
\clause{\boxvar{b2} \concept{be} "v.03" \disvar{e1}}    & \clause{\boxvar{b2} \role{Location} \disvar{e1} \disvar{x1}} \\ 
\multicolumn{2}{l}{\clause{\boxvar{b2} \role{YearOfCentury} \disvar{t2} "2013"}}  \\
\bottomrule
\end{tabular}
\end{minipage}%
\begin{minipage}[t]{.75\linewidth}
\vspace{0.01cm}
\centering
\includegraphics[width=0.66\textwidth,right]{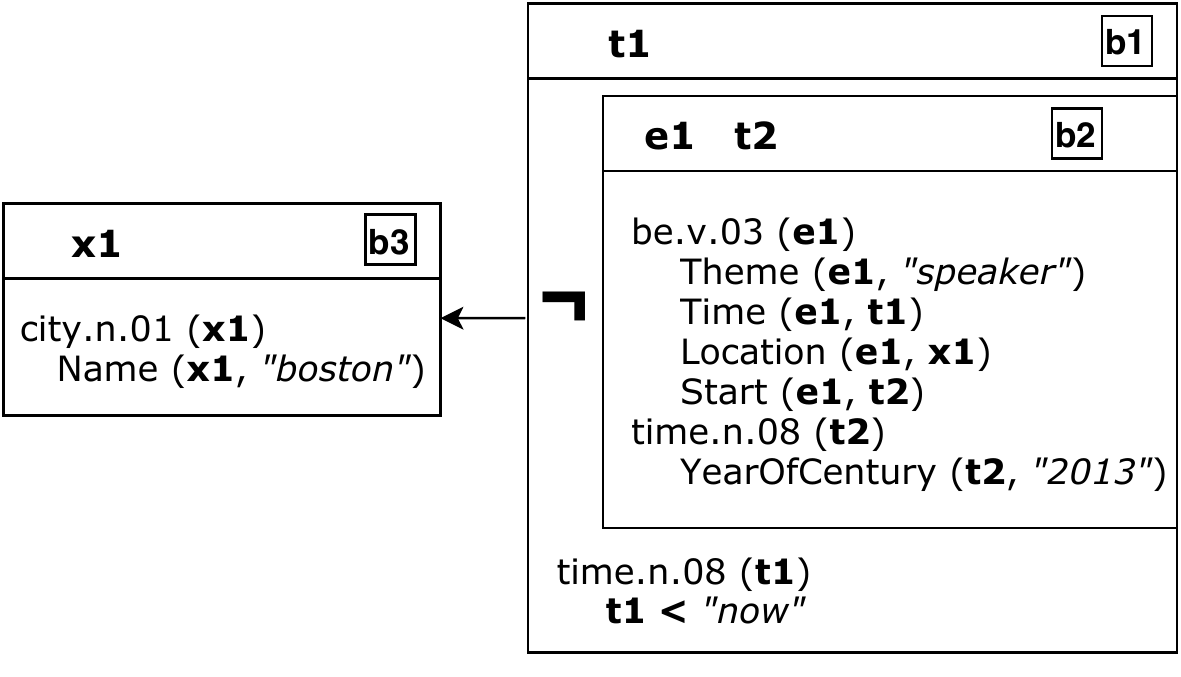}
\end{minipage}%

\vspace{-0.3cm}
\caption{\label{fig:clauses}\label{fig:drs}Example DRS in both clause (left) and box (right) representation.}
\vspace{-0.3cm}
\end{figure*}

\inlineheader{Why Discourse Representation Structures?} DRS parsing is a task that combines logical, pragmatic and lexical components of semantics in a single meaning representation. The task is complex and comprises other NLP tasks, such as semantic role labeling, word sense disambiguation, co-reference resolution and named entity tagging. Also, DRSs show explicit scope for certain operators, which allows for a more principled and linguistically motivated treatment of negation, modals and quantification, as has been advocated in formal semantics.
Moreover, DRSs can be translated to formal logic, which allows for automatic forms of inference by third parties. Lastly, annotated DRSs are available in four languages  (\citealp{eacl:pmb}, see Section~\ref{sec:data}), allowing us to evaluate our models on multiple languages.

\section{Background}
\label{sec:background}

\subsection{Character-level models}
\label{sec:backgroundchar}

The power of character-level representations has long been known in the field. 
In earlier work, they were successfully used in a range of tasks, including
text-to-speech \citep{sejnowski1987parallel}, parallel text alignment \citep{church1993char_align}, grapheme to phoneme conversion \citep{kaplan:94}, language identification \citep{dunning1994statistical}, topical similarity prediction \citep{cavnar1995using}, named entity recognition \citep{klein2003named}, authorship attribution \citep{peng2003language} and statistical machine translation \citep{vilar2007can}.

More recently, they also proved useful as input representations for neural networks, starting with success in general language modelling \citep{sutskever:11,kim2016character,bojanowski2017enriching}, but also for a range of other tasks, including tokenization \citep{evang2013elephant}, POS-tagging \citep{santos2014learning, plank:16}, dependency parsing \citep{ballesteros-etal-2015-improved, vania-etal-2018-character} and neural machine translation \citep{chung-etal-2016-character,costa2016character, luong-manning-2016-achieving, cherry-etal-2018-revisiting}.

In semantic parsing, if character-level representations are employed, they are commonly used in combination with non-contextual word-level representations \citep{lewis-etal-2016-lstm, ballesteros2017amr,groschwitz-etal-2018-amr, cai-lam-2019-core}. There are a few recent studies that did use character-level representations in combination with \bert{} \citep{zhang-etal-2019-amr,zhang-etal-2019-broad,cai-lam-2020-amr}, though only \citet{zhang-etal-2019-amr} provided an ablation score without the characters. Moreover, it is not clear if this small improvement was significant.
\citet{clinAMR:17} and \citet{drstacl:18}, on the other hand, used solely character-level representations in an end-to-end fashion, using a bi-LSTM sequence-to-sequence model, which outperformed word-based models that employed non-contextual embeddings.

\subsection{Discourse Representation Structures}
\label{sec:backgrounddrs}

DRSs are formal meaning representations introduced by Discourse Representation Theory \citep{kampreyle:drt} with the aim to capture the meaning of texts (Figure~\ref{fig:drs}).
Many variants of DRS have been proposed throughout the years.
We adopt 
\citet{venhuizen2018discourse}'s version of DRT, which is close to Kamp's original ideas, but has 
a neo-Davidsonian view of event semantics and explicitly represents presuppositions.

\inlineheader{Corpora} The Groningen Meaning Bank (GMB, \citealp{gmb:eacl,GMB:2017}) was the first attempt of annotating open domain English texts with DRSs. The released documents are partially corrected, but there are no gold standard sets available for evaluation. A similar corpus is the Parallel Meaning Bank (PMB, \citealp{eacl:pmb}), which builds upon the GMB in a number of ways. It contains (parallel) texts in four languages: English, German, Italian and Dutch, with more fine-grained and language-neutral DRSs. Semantic tags are used during annotation \citep{Bjervaetal:16, semantic-tagset:17}, and all non-logical DRS symbols are  grounded in either WordNet \citep{wordnet} or VerbNet \citep{Bonial:11}. Moreover, its releases contain gold standard DRSs. For these reasons, we take the PMB as our corpus of choice to evaluate our DRS parsers.

\inlineheader{DRS parsing} Early approaches to DRS parsing employed rule-based systems for small English texts \citep{johnsonklein, wadaasher, Bos2001ICOS}. The first open domain DRS parser is Boxer \citep{step2008:boxer, boxer}, which is a combination of rule-based and statistical models. 
\citet{le:12} used a probabilistic parsing model that used dependency structures to parse GMB data as graphs. More recently, \citet{neural_drs_gmb:18} proposed a neural model that produces (tree-structured) DRSs in three steps by first learning the general (box) structure of a DRS, after which specific conditions and referents are filled in. In follow-up work \citep{liu-etal-2019-acl} they extend this work by adding an improved attention mechanism and constraining the decoder to ensure well-formed output. This model achieved impressive performance on both sentence-level and document-level DRS parsing on GMB data. \citet{fu2020drts} in turn improve on this work by employing a Graph Attention Network during both encoding and decoding.

The introduction of gold standard DRSs in the PMB enabled a principled comparison of approaches. 
In our previous work \citep{drstacl:18}, we showed that sequence-to-sequence models can successfully learn to produce DRSs, with characters as the preferred representation. In follow-up work, we improved on these scores by adding linguistic features \citep{van-noord-etal-2019-linguistic}. The first shared task on DRS parsing \citep{abzianidze-etal-2019-first} sparked more interested in the topic, with a system based on stack-LSTMs \citep{evang-2019-transition} and a neural graph-based system \citep{fancellu-etal-2019-semantic}. The best system \citep{liu-etal-2019-discourse} used a similar approach as \citet{drstacl:18}, but swapped the bi-LSTM encoder for a Transformer. We will compare our approach to these models in Section~\ref{sec:results}.

\vspace{-0.1cm}
\section{Method}
\vspace{-0.1cm}

\subsection{Neural Architecture}
\label{sec:neural}

As our baseline system, we start from a fairly standard sequence-to-sequence model with attention \citep{bahdanau2015neural}, implemented in AllenNLP \cite{Gardner2017AllenNLP}.\footnote{\url{https://github.com/RikVN/allennlp}} We improve on this model in a number of ways, mainly based on Nematus \citep{sennrich-etal-2017-nematus}: (i) we initialize the decoder hidden state with the mean of all encoder states, (ii) we add an extra linear layer between this mean encoder state and the initial decoder state and (iii) we add an extra linear layer after each decoder state.

Specifically, given a source sequence $(s_1,\dots, s_l)$ of length $l$, and a target sequence $(t_1,\dots, t_k)$ of length $k$, let $\vect{e}_i$ be the embedding of source symbol $i$, let $\vect{h}_i$ be the encoder hidden state at source position $i$ and let $\vect{d}_j$ be the decoder state at target position $j$. A single forward encoder state is obtained as follows:  $\overrightarrow{\vect{h}}_i = \text{LSTM} (\overrightarrow{\vect{h}}_{i-1}, \vect{e}_i)$. The final state is obtained by concatenating the forward and backward hidden states,  $\vect{h}_i = [ \overrightarrow{\vect{h}}_i; \overleftarrow{\vect{h}}_i ]$. The decoder is initialized with the average over all encoder states:  $ \vect{c}_\mathrm{tok} = \left(\sum_{i=1}^{l}\vect{h}_i \right) / \: l$ and $\vect{d}_0 = \tanh \left( \matr{W}_\mathrm{init}\:  \vect{c}_\mathrm{tok} \right)$.

\inlineheader{Characters in one encoder} We will experiment with adding character-level information in either one or two encoders. For one encoder, we use char-CNN \citep{kim2016character}, which runs a Convolutional Neural Network \citep{lecuncnn:90} over the characters for each token. It applies convolution layers for certain \emph{widths}, which in essence select n-grams of characters. For each width, it does this a predefined number of times, referred to as the number of \emph{filters}. The filter vectors form a matrix, which is then pooled to a vector by taking the max value of each initial filter vector. A detailed schematic overview of this procedure is shown in Appendix~\ref{app:charcnn}. However, we usually do not look at only a single width, but at a range of widths, e.g., $\left[ 1, 2, 3, 4, 5 \right]$. In that case, we simply concatenate the resulting vectors to obtain our final char-CNN embedding: $\vect{e}_\mathrm{char_i} = \left[ \vect{e}_{w1} ; \vect{e}_{w2} ; \vect{e}_{w3} ; \vect{e}_{w4} ; \vect{e}_{w5} \right]$. Each width-filter combination has independent learnable parameters. Finally, the char-CNN embedding is concatenated to the token-level representation, which is fed to the encoder: $\vect{e}_i = [ \vect{e}_\mathrm{tok_i}; \vect{e}_\mathrm{char_i}  ]$.

\begin{figure}[!htb]
  \includegraphics[width=0.5\textwidth]{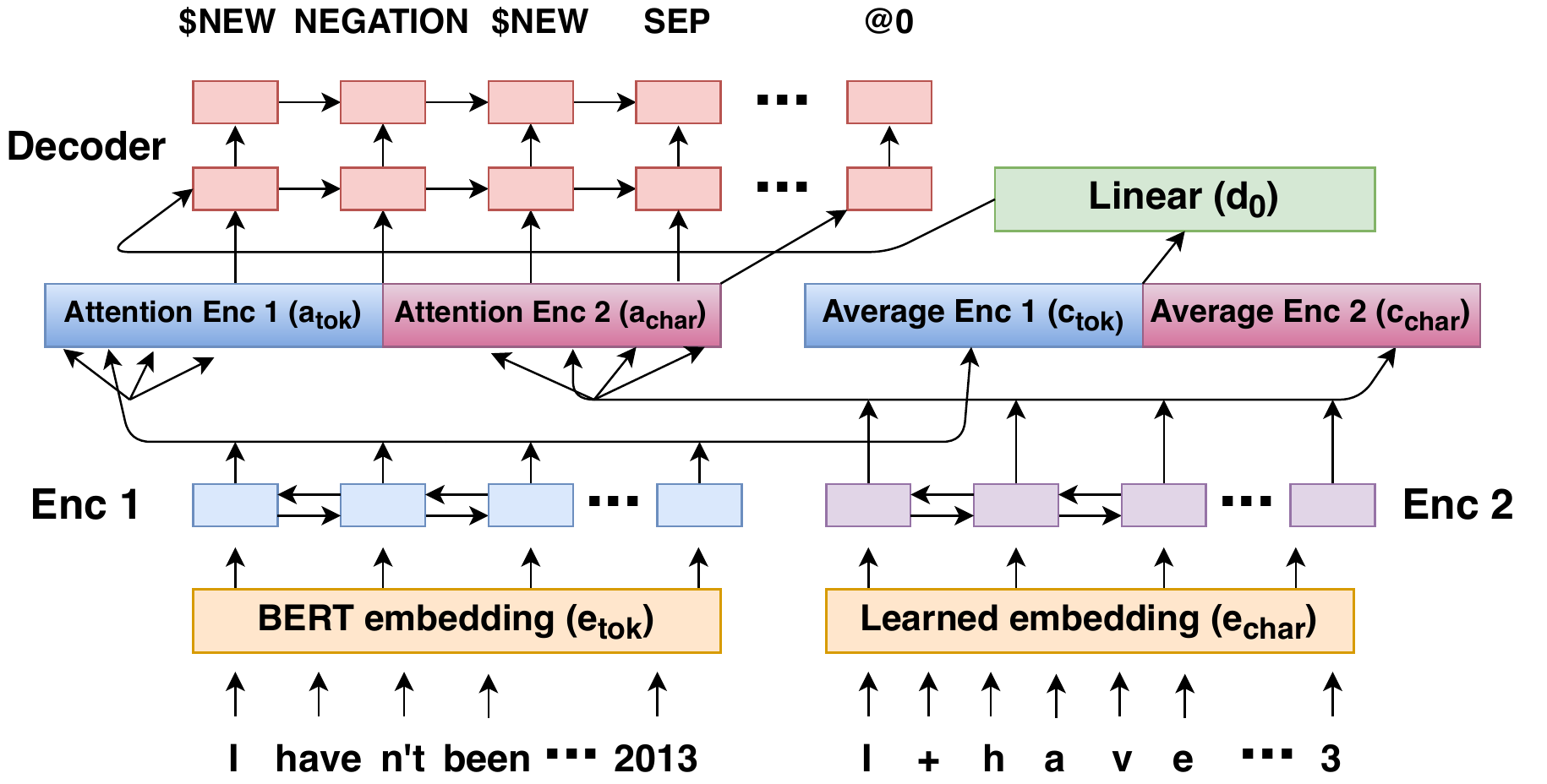}
  \caption{Schematic overview of our neural architecture when using two encoders (\bert{} and characters).\label{fig:model}}
\end{figure}

\inlineheader{Characters in two encoders} In the two-encoder setup, we run separate (but structurally identical) bi-LSTM encoders over the tokens and characters, and concatenate the resulting context vector before we feed it to the decoder: 

\begin{equation*}
    \vect{d}_0 = \tanh \left( \matr{W}_\mathrm{init} \left[ \vect{c}_\mathrm{tok}; \vect{c}_\mathrm{char} \right] \right)
\end{equation*}

In the decoder, we replace the LSTM with a doubly-attentive LSTM, based on the doubly-attentive GRU \citep{calixto-etal-2017-doubly}. We apply soft-dual attention \citep{junczys-dowmunt-grundkiewicz-2017-exploration} to be able to attend over both encoders in the decoder (also see Figure~\ref{fig:model}):

\vspace{-0.4cm}
\begin{alignat*}{2}
    &\vect{d}_j^{\prime} &&= \text{LSTM}_1 \left( \vect{d}_{j-1}, \vect{e}_{t_{j-1}}  \right) \\
    &\vect{a}_j &&= \left[ \text{ATT}\left( \matr{C}_\mathrm{tok},\vect{d}_j^{\prime}   \right); \text{ATT}\left( \matr{C}_\mathrm{char},\vect{d}_j^{\prime}   \right) \right] \\
    &\vect{d}_j &&= \text{LSTM}_2 \left( \vect{d}^{\prime}_j, \vect{a}_j   \right)
\end{alignat*}
\vspace{-0.4cm}

Here, $\vect{e}_{t_{j-1}}$ is the embedding of the previously decoded symbol $t$,  $C$ the set of encoder hidden states for either the tokens or characters, ATT the attention function (dot-product) and $\vect{d}_j$ the final decoder hidden state at step $j$. This model can easily be extended to more than two encoders, which we will experiment with in Section~\ref{sec:results}.

This type of multi-source model is commonly used to represent different languages, e.g., in machine translation \citep{zoph:16,firat-etal-2016-multi} and semantic parsing \citep{susanto2017neural, duong2017multilingual}, though it has also been successfully applied in multi-modal translation \citep{libovicky-helcl-2017-attention}, multi-framework semantic parsing \citep{stanovsky-dagan-2018-semantics} and adding linguistic information \citep{currey:18, van-noord-etal-2019-linguistic}. To the best of our knowledge, we are the first to represent the characters as a source of extra information in a multi-source sequence-to-sequence model.

\inlineheader{Transformer} We also experiment with the Transformer model \citep{transformer:17}, using the stacked self attention model as implemented in AllenNLP. A possible advantage of this model is that it might handle longer sentences and documents better. However, it might be harder to tune \citep{popel2018training}\footnote{Also see: \url{https://twitter.com/srush_nlp/status/1245825437240102913}} and its improved performance has mainly been shown for large data sets, as opposed to the generally smaller semantic parsing data sets (Section~\ref{sec:data}). Indeed, we cannot outperform the LSTM architecture (see Section~\ref{sec:results}), even when tuning more extensively. We therefore do not experiment with adding character-level representations to this architecture, though the char-CNN can be added similarly as for the LSTM model.

\inlineheader{Hyper-parameters} To make a fair comparison, we conduct an independent hyper-parameter search on the development set for all nine input text representations (see Section~\ref{sec:lms}) across the two neural architectures, starting from the settings of \citet{van-noord-etal-2019-linguistic}. We found that the best settings were very close for all systems, with the only notable difference that the learning rate of the Transformer models is considerably smaller than for the bi-LSTM models ($0.0002$ vs $0.001$).\footnote{See Appendix~\ref{sec:hyperappendix} for specific hyperparameter settings.}

For the char-CNN model, we use $100$ filters, an embedding size of $75$ and n-gram filter sizes of $[1, 2, 3]$ for English and $[1, 2, 3, 4, 5]$ for German, Italian and Dutch. For experiments where we add characters or linguistic features, the only extra search we do is the size of the hidden vector of the RNN encoder ($300 - 600$), since this vector now has to contain more information, and could potentially benefit from a larger size. Note that (possible) improved performance is not simply due to larger model capacity, since during tuning of the baseline models a larger RNN hidden size did not result in better performance.

\subsection{Representations}
\label{sec:lms}

We will experiment with five well-known pretrained language models: \elmo{} \citep{Peters:2018}, \bert{} base/large \citep{bert:19} and \roberta{} base/large \citep{roberta_arxiv:19}.\footnote{We are aware that there exist several other large pre-trained language models (e.g., \citealp{yang2019xlnet,2019t5,clark2020electra}), but we believe that the models we used have had the largest impact on the field.} The performance of these five large LMs is contrasted with results of a character-level model and three word-based models. The word-based models either learn the embeddings from scratch or use non-contextual \glove{} \citep{glove:14} or \fasttext{} \citep{grave2018learning} embeddings. 
Pre- and postprocessing of the DRSs is done using the method described in \citet{drstacl:18}.\footnote{\url{https://github.com/RikVN/Neural_DRS/}} The DRSs are linearized, after which the variables are rewritten to a relative representation. The character-level model has character representations for the DRS concepts and constants, but not for variables, roles and operators. For all word-level models, the DRS concepts are initialized with \glove{} embeddings, while the other target tokens are learned from scratch.

\inlineheader{BERT specifics} For the \bert{} models, we obtained the best performance by only keeping the vector of the first WordPiece per original token (e.g., only keep \texttt{play} out of \texttt{play \#\#ing}). For \roberta{}, it was best to use the WordPiece tokenization as is. Since linguistic features are added on token level, we duplicate the semantic tags for multi-piece tokens of \roberta{} in Table~\ref{tab:lmcomparison}. Interestingly, we found that for both \bert{} and \roberta{}, it was best to keep the pretrained weights frozen. This was not a small difference: models using fine-tuning always obtained low scores ($45$ to $60$).

\subsection{Data and Evaluation}
\label{sec:data}

We use PMB releases 2.2.0 and 3.0.0\footnote{\url{https://pmb.let.rug.nl/data.php}} in our experiments (Table~\ref{tab:releases}). The latter is a larger and more diverse extension of 2.2.0, which will be used for most of our experiments. We use 2.2.0 to compare to previous work and to verify that our results are robust across datasets. The PMB releases contain DRSs for four languages (English, German, Italian and Dutch) for three levels of annotation: gold (fully manually checked), silver (partially manually corrected) and bronze (no manual corrections). To make a fair comparison to previous work, we only employ the gold and silver data, by pretraining on gold $+$ silver data and subsequently fine-tuning on only the gold data. If there is no gold train data available, we train on silver $+$ bronze and fine-tune on silver. Unless otherwise indicated, our results are on the English development set of release 3.0.0.

\begin{table}[!t]
\centering
\setlength{\tabcolsep}{4pt}
\resizebox{0.95\columnwidth}{!}{
\begin{tabular}{llrrrrr}
\toprule 
& & \multicolumn{3}{c}{\textbf{Gold}} & \multicolumn{1}{c}{\textbf{Silver}} & \multicolumn{1}{c}{\textbf{Bronze}} \\
& & \multicolumn{1}{c}{Train} & \multicolumn{1}{c}{Dev} & \multicolumn{1}{c}{Test} & \multicolumn{1}{c}{Train} & \multicolumn{1}{c}{Train} \\
 \midrule
\bf 2.2.0 & English & 4,597 & 682 & 650 & 67,965 & 120,662 \\
& German & 0 & 727 & 747 & 4,235 & 102,998 \\
& Italian & 0 & 374 & 400 & 2,515 & 61,504 \\
& Dutch & 0 & 370 & 341 & 1,051 & 20,554 \\
\midrule
\bf 3.0.0 & English & 6,620 & 885 & 898 & 97,598 & 146,371\\
& German & 1,159 & 417 & 403 & 5,250 & 121,111 \\
& Italian & 0 & 515 & 547 & 2,772 & 64,305 \\
& Dutch & 0 & 529 & 483 & 1,301 & 21,550\\
\bottomrule
\end{tabular}
}
\caption{Number of documents for the four languages, for the two PMB releases considered.\label{tab:releases}}
\end{table}

\inlineheader{Linguistic features} We want to contrast our method of character-level information to adding sources of linguistic information. Based on \citet{van-noord-etal-2019-linguistic}, we employ these five sources: part-of-speech tags (POS), dependency parses (DEP), lemmas (LEM), CCG supertags (CCG) and semantic tags (SEM). For the first three sources, we use Stanford CoreNLP \citep{manning2014stanford} to parse the documents in our dataset. The CCG supertags are obtained by using easyCCG \citep{lewisSteedman:14}. For semantic tagging, we train our own trigram-based tagger using TnT \citep{Brants:2000}.\footnote{This tagger is also used in the PMB pipeline, see \citet{semantic-tagset:17}. It outperformed an ngram-based CRF-tagger \citep{crf:01} we also tried, obtaining an accuracy of $94.4\%$ on the dev set.} Table~\ref{tab:ling} shows a tagged example sentence for all five sources of information. Moreover, we also include non-contextual \glove{} and \fasttext{} embeddings as an extra source of information.

We add these sources of linguistic information in the same way as we add the character-level information, in either one or two encoders (see Section ~\ref{sec:neural}). In two encoders, we can use the exact same architecture. For one encoder, we (obviously) do not use the char-CNN, but learn a separate embedding for the tags (of size 200), that is then concatenated to the token-level representation, i.e., $\vect{e}_i = [ \vect{e}_\mathrm{tok_i}; \vect{e}_\mathrm{ling_i}  ]$. If we use two encoders with a LM, characters \emph{and} linguistic information (e.g., Table~\ref{tab:charplusling}), the characters are added separately in the second encoder, while the LM and linguistic information representations are added in the first encoder.

\begin{table}[!tb]
\centering
\renewcommand{\arraystretch}{1.2}
\setlength{\tabcolsep}{2pt}
\resizebox{\columnwidth}{!}{

\begin{tabu}{lllllllll}
\toprule
\textbf{Sent} & I & have & n't & been & to & Boston & since & 2013 \\
\midrule
\textbf{POS} & \postag{PRP} & \postag{VBP} & \postag{RB} & \postag{VBN} & \postag{TO} & \postag{NNP} & \postag{IN} & \postag{CD} \\
\textbf{SEM} & \semtag{PRO} & \semtag{NOW} & \semtag{NOT} & \semtag{EXT} & \semtag{REL} & \semtag{GPE} & \semtag{REL} & \semtag{YOC} \\
\textbf{LEM} & \lemtag{I} & \lemtag{have}  & \lemtag{not}  & \lemtag{be}  & \lemtag{to}  & \lemtag{Boston}  & \lemtag{since}  & \lemtag{2013} \\
\textbf{DEP} & \deptag{nsubj} & \deptag{aux} & \deptag{neg} & \deptag{cop} & \deptag{case} & \deptag{ROOT} & \deptag{case} & \deptag{nmod} \\
\textbf{CCG}  & \ccgtag{NP}    & \ccgtag{VP\textbackslash{}VP} & \ccgtag{VPVP} & \ccgtag{VP/PP} & \ccgtag{PP/NP} & \ccgtag{N} & \ccgtag{\small(VP\textbackslash{}VP)/NP} & \ccgtag{N}   \\
\bottomrule
\end{tabu}
}
\caption{\label{tab:ling}Example representation for each source of linguistic information (PMB document p00/d1489).}
\vspace*{-0.3cm}
\end{table}

\inlineheader{Evaluation} We compare the produced DRSs to the gold standard using Counter \citep{pmb-LREC:18}, which calculates micro precision, recall and F1-score based on the number of matching clauses.\footnote{\url{https://github.com/RikVN/DRS_parsing/}} We use Referee \citep{drstacl:18} to ensure that the produced DRSs are syntactically and semantically well-formed (i.e., no free variables, no loops in subordinate relations) and form a connected graph. DRSs that are ill-formed get an F1-score of $0.0$. All shown scores are averaged F1-scores over five training runs of the system, in which the same five random seeds are used.\footnote{Standard deviations are omitted for brevity, though available for all experiments here: \url{https://github.com/RikVN/Neural_DRS/}} For significance testing we use approximate randomization \citep{random-appr:89}, with $\alpha = 0.05$ and $R = 1000$.

We also introduce and release \drseval{}. This program provides a detailed overview of the performance of a DRS parser, but can also compare experiments, possibly over multiple runs. Features include significance testing, semantic tag analysis (Section~\ref{sec:semanalysis}), sentence length plotting (Section~\ref{sec:lenanalysis}), new detailed Counter scores (Appendix~\ref{app:detailed}), and analysing (relative) best/worst produced DRSs (Appendix~\ref{app:sents}). We hope this is a step in the direction of a more principled way of evaluating DRS parsers. 

\vspace{-0.1cm}
\section{Results}
\label{sec:results}
\vspace{-0.1cm}

\inlineheader{LMs vs char-level models} DRS parsing is no exception to the general trend in NLP: it is indeed the case that the pretrained language models outperform the char-only model (Table~\ref{tab:charvsbert}). Interestingly, the Transformer model has worse performance for all representations.\footnote{The Transformer models were even tuned longer, since they were more sensitive to small hyperparameter changes.} Surprisingly, we find that \bertbase{} is the best model, though the differences are small.\footnote{\bertbase{} significantly outperformed all the other models, except for \bertlarge.} We use this model in further experiments (referred to as \bert{}).

\inlineheader{Adding characters to BERT} We can see the impact of adding characters to \bert{} (first row of results in Table~\ref{tab:charplusling}). For both methods, it results in a clear and significant improvement over the \bert{}-only baseline, 87.6 versus 88.1. 

\begin{table}[!tb]
\centering
\setlength{\tabcolsep}{3pt}
\resizebox{0.875\columnwidth}{!}{
\begin{tabular}{lcc}
\toprule
                             & \textbf{bi-LSTM} & \textbf{Transformer} \\
 \midrule  
\bf Char & 86.1 & 79.7 \\
\bf Word & 85.3 & 83.6\\
\bf  \glove{} & 85.4 & 84.6 \\
\bf \fasttext{} & 85.5 & 84.0  \\
\bf  \elmo{} & 87.3  & 84.3   \\
\bf  \bertbase{} & \bf 87.6 & \bf 85.4 \\
\bf  \bertlarge{} & 87.5  & 84.7     \\
\bf  \robertabase{} & 87.0  & 82.7    \\
\bf  \robertalarge{} & 86.8  & 81.9  \\
\bottomrule
\end{tabular}
}
\caption{\label{tab:charvsbert}Baseline model for the nine input representations considered, for the bi-LSTM and Transformer architectures. Best score in each column shown in bold.}
\end{table}

\begin{table}[!tb]
\centering
\setlength{\tabcolsep}{4pt}
\resizebox{\columnwidth}{!}{
\begin{tabular}{l|cc|ccc}
\toprule
  & \multicolumn{2}{c|}{\textbf{No chars}} & \multicolumn{3}{c}{\textbf{+ characters}} \\
  & \textbf{1-enc} & \textbf{2-enc} & \textbf{1-enc} & \textbf{2-enc} & \textbf{3-enc} \\
\midrule                             
\textbf{\bert}            & 87.6   &  NA      &  \bf 88.1   &   \bf 88.1    & NA        \\
\midrule
\textbf{\bert{} + word} & 87.7 & 87.4 & 87.8 & 87.6 & 86.9\\
\textbf{\bert{} + \glove{}}    & 87.9   & 87.2    &  88.1       & 88.0 & 86.9  \\
\textbf{\bert{} + \fasttext{}} & 87.8   & 87.7    &  87.9      & 87.9  & 87.0   \\
\textbf{\bert{} + pos}      & 87.6   & 87.6       &  87.4      & 87.6  & 87.8 \\
\textbf{\bert{} + sem}      & 87.9   & 88.0       &  88.0      & \bf 88.4 & 88.1  \\
\textbf{\bert{} + lem}      & 87.8   & 88.0       &  88.1      & 88.0  & 87.4 \\
\textbf{\bert{} + dep}      & 87.9   & 87.5       &  88.0       & 87.8  & 87.8  \\
\textbf{\bert{} + ccg}      & 87.8   & 87.3       & 87.9     & 87.8  & 87.6  \\ 
\bottomrule
\end{tabular}
}
\caption{Results for adding characters, linguistic information and a combination of the two to the bi-LSTM \bertbase{} model on 3.0.0 English dev.\label{tab:charplusling}}
\vspace{-0.3cm}
\end{table}

\inlineheader{Adding linguistic features to BERT} However, another common method of improving performance is adding linguistic features to the token-level representations. We try a range of linguistic features (described in Section~\ref{sec:data}), that are added in either one or two encoders. We see in the first two columns of results of Table~\ref{tab:charplusling} that even though linguistic information sources indeed do improve performance (up to $0.4$ absolute), there is no single source that can beat adding just the character-level representations ($88.1$).

\inlineheader{Combining characters and linguistic features} An obvious follow-up question is whether we still see improvements for character-level models when also adding linguistic information. 
In a single encoder, adding characters (third column of results in Table~\ref{tab:charplusling}) is beneficial for 6 out of 7 linguistic sources (i.e., compared to the first column of results). The scores are, however, not higher than simply adding characters on their own, suggesting that linguistic features are not always beneficial if character-level features are also included. For two encoders, the pattern is less clear, but we do find our highest score thus far when we combine characters and semantic tags (88.4).\footnote{This improvement is significant. With gold semantic tags (ceiling performance) we score $88.6$.} Using three encoders did not yield clear improvements over two encoders. Therefore, we do not experiment with using more than three encoders.

\begin{table}[!tb]
\centering
\setlength{\tabcolsep}{2pt}
\resizebox{\columnwidth}{!}{
\begin{tabular}{lcccc}
\toprule
                       & \textbf{No char} & \textbf{\begin{tabular}[c]{@{}l@{}}+char\\ (1 enc)\end{tabular}} & \textbf{\begin{tabular}[c]{@{}l@{}}+char\\ (2 enc)\end{tabular}} & \textbf{\begin{tabular}[c]{@{}c@{}}+char +sem\\       (2 enc)\end{tabular}} \\
\midrule
\textbf{\elmo}         & 87.3   & 87.6    & 87.8    & \bf 88.0      \\
\textbf{\bertbase}     & 87.6   & 88.1    & 88.1    & \bf 88.4        \\
\textbf{\bertlarge}    & 87.5   & 88.2    & 87.8    & \bf 88.3       \\
\textbf{\robertabase}  & 87.0   & 87.3    & 87.8    & \bf 88.0       \\
\textbf{\robertalarge} & 86.8   & 86.8    & 87.0    & \bf 87.3       \\
\bottomrule
\end{tabular}
}
\caption{Results on 3.0.0 English dev of four LMs for adding characters and both characters and semtags.\label{tab:lmcomparison}}
\vspace{-0.3cm}
\end{table}

\begin{figure*}[!htb]
\centering
\begin{subfigure}{.5\textwidth}
  \centering
  \includegraphics[width=\columnwidth]{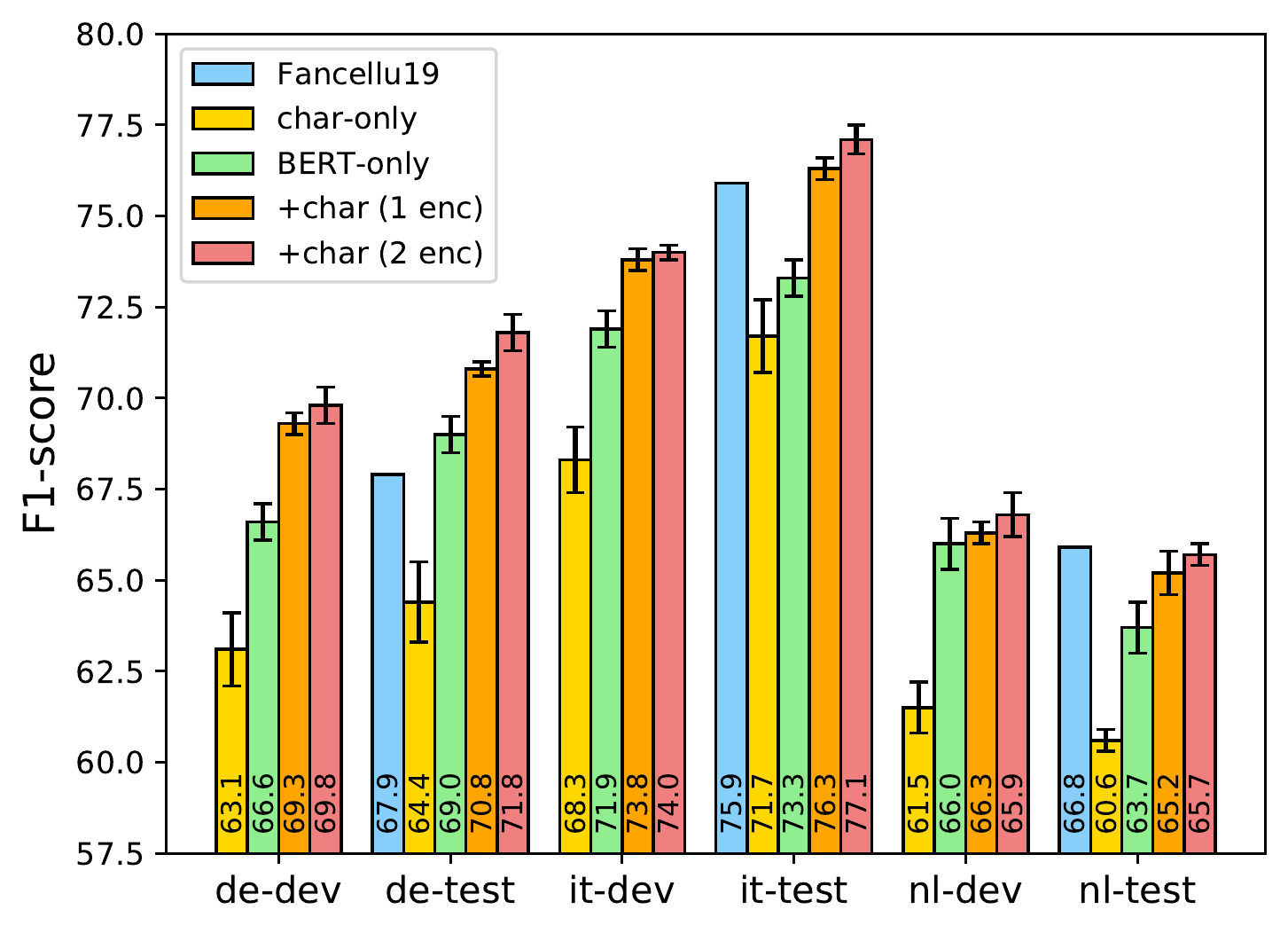}
  \caption{Scores for PMB release 2.2.0.}
\end{subfigure}%
\begin{subfigure}{.5\textwidth}
  \centering
 \includegraphics[width=\columnwidth]{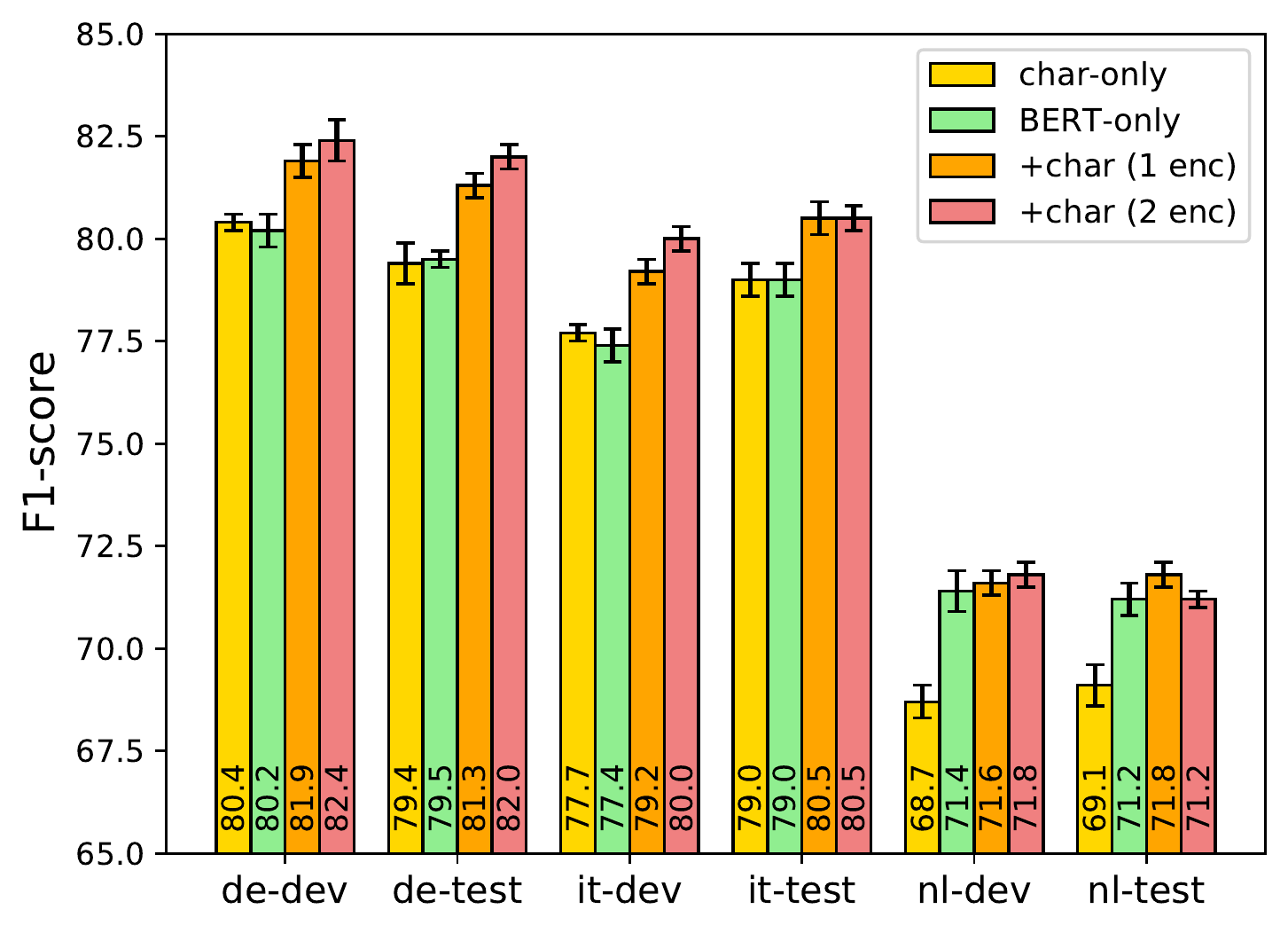}
   \caption{Scores for PMB release 3.0.0.}
  \label{fig:sub2}
\end{subfigure}
\caption{\label{fig:nonen}Dev and test scores (F1) for the four models we trained for three languages (German, Italian and Dutch). For 2.2.0, we compare our results to \citet{fancellu-etal-2019-semantic}.}
\vspace{-0.3cm}
\end{figure*}

\inlineheader{Robustness to different LMs} We want to verify that the character improvements are robust to using different language models (Table~\ref{tab:lmcomparison}). We see that adding characters results in improvement for all the LMs under consideration, even for \elmo{}, which already incorporates characters in creating the initial embeddings. Moreover, combining characters and semantic tags also results in an improvement over just using characters for all the LMs considered.

\inlineheader{Robustness across languages} We train systems for German, Italian and Dutch for four models: char-only, \bertonly{}, \bert{} + char in 1 encoder, and \bert{} + char in two encoders.\footnote{We do not train a model that uses semantic tags as features, since there is not enough gold semantic tag data available to train a good tagger for any of these languages.} The \bert{} model we use is \texttt{bert-multilingual-uncased}. The results for both PMB releases are shown in Figure~\ref{fig:nonen}. For all languages, adding characters leads to a clear improvement for both one and two encoders, though for Dutch the improvement is smaller than for German and Italian. Interestingly, the two-encoder setup seems to be preferable for these smaller, non-English data sets. For 2.2.0, we outperform the system of \citet{fancellu-etal-2019-semantic} for German and Italian and obtain competitive scores for Dutch.

\begin{table}[!htb]
\setlength{\tabcolsep}{4pt}
\resizebox{\columnwidth}{!}{
\begin{tabular}{l|rr|rr}
\toprule
  & \multicolumn{2}{c|}{\textbf{2.2.0}} & \multicolumn{2}{c}{\textbf{3.0.0}} \\
  & \textbf{Dev} & \textbf{Test} & \textbf{Dev} & \textbf{Test} \\
\midrule
\textbf{Amateur Boxer}  & 72.2 & 72.2 & 78.2  & 78.8 \\
\textbf{Pro Boxer}  & NA & NA & 88.2  & 88.9 \\
\textbf{\citet{fancellu-etal-2019-semantic}} & NA & 76.4 & NA & NA \\
\textbf{\citet{evang-2019-transition}}  & 74.4 & 74.4 & NA & NA \\
\textbf{\citet{drstacl:18}} & 81.2 & 83.3 & 84.3 & 84.9  \\
\textbf{\citet{van-noord-etal-2019-linguistic}} & 86.5 & 86.8 & 86.8  & 87.7  \\
\textbf{\citet{liu-etal-2019-discourse}} & 85.5 & 87.1 & NA & NA \\
\midrule
\textbf{This work - \bert{}} & 85.4 & 87.9 & 87.6 & 88.5 \\
\textbf{This work - \bert{} + char (1 enc)} & \bf 86.1 & \bf 88.3 & 88.1 & 89.2  \\
\textbf{This work - \bert{} + char (2 enc)} & 85.6 & 88.1 & 88.1 & 89.0 \\
\textbf{This work - Best model} & 85.5 & 87.7 & \bf 88.4 & \bf 89.3 \\
\bottomrule
\end{tabular}
}
\caption{Comparison of our four main models to previous work for PMB 2.2.0 and 3.0.0 (English only).\label{tab:finalres}}
\vspace{-0.3cm}
\end{table}

\inlineheader{Comparison to previous work} To check whether the improvements hold on unseen data, we run our best models on the test set and compare the scores to previous work (Table~\ref{tab:finalres}).\footnote{For the detailed Counter scores see Appendix~\ref{app:detailed}.} 
We see that adding the character-level information has similar (significant) improvements for dev and test on both data sets. The addition of semantic tags might be questionable: for 2.2.0, both the \bert{} + char models outperform this model, while for 3.0.0 the 0.1 improvement over \bert{} + char in one encoder is not significant. Despite this, we reach state-of-the-art performance on both data sets, significantly outperforming the previous best scores by \citet{van-noord-etal-2019-linguistic} and \citet{liu-etal-2019-discourse}. We also compare to the semantic parser Boxer, which needs input for 6 different PMB layers \citep{eacl:pmb}. \emph{Amateur Boxer} is trained with internal PMB taggers, while \emph{Pro Boxer} uses the output of a neural multi-task learning system based on \bert{} \citep{machamp:20}. Even though this is an unfair comparison to our system, since the rule-based components of Boxer are (partly) optimized on the dev and test sets, our best model still improves slightly over \emph{Pro Boxer} (significantly on test).

\section{Analysis}
\label{sec:analysis}

\begin{table}[!t]
\centering
\setlength{\tabcolsep}{1pt}
\resizebox{\columnwidth}{!}{
\begin{tabular}{lrcccc}
\toprule
& \textbf{\# Docs} & \textbf{\bert{}} & \textbf{+char} & \textbf{+char} & \textbf{+ch+sem} \\
& & & \bf (1 enc) & \bf (2 enc) & \bf (2 enc) \\
\midrule
\bf All sentences & 1,783 & 88.1 & 88.7 & 88.5 & \bf 88.8 \\
\midrule  
\bf Modality            & 188 & 86.8 & \cellgreen +0.1 & \cellgreen +0.1 & \cellgreen \bf +0.4 \\
\bf \quad Negation    & 156 & 88.8 & \cellgreen +0.2 & \cellred -0.1 & \cellgreen \bf +0.4 \\
\bf \quad Possibility & 38  & 81.3 & \cellorange 0.0 & \cellgreen +1.0 & \cellgreen \bf +1.5 \\
\bf \quad Necessity   & 13  & 74.5 & \cellred -1.6 & \cellgreen \bf +1.4 & \cellred -0.2 \\
\bf Logical & 449 & 86.3 & \cellgreen \bf +0.7 & \cellgreen +0.2 & \cellgreen +0.5 \\
\bf Pronouns & 996 & 88.9 & \cellgreen +0.4 & \cellgreen +0.4 & \cellgreen \bf +0.6 \\
\bf Attributes & 1,063 & 87.6 & \cellgreen +0.7 & \cellgreen +0.4 & \cellgreen \bf +0.8 \\
\bf Comparatives & 45 & 84.5 & \cellgreen \bf +1.6 & \cellgreen +0.2 & \cellred -0.2 \\
\bf Named entities & 673 & 88.1 & \cellgreen +0.5 & \cellgreen +0.3 & \cellgreen \bf +0.6 \\
\bf Numerals & 186 & 85.8 & \cellgreen +1.1 & \cellgreen +1.2 & \cellgreen \bf +1.5 \\
\bottomrule
\end{tabular}
}
\caption{\label{tab:semtag}F-scores on subsets of sentences that contain a certain phenomenon, based on semantic tags, for the combined dev and test set of PMB release 3.0.0. Full scores shown for \bert{} and absolute differences for the remaining systems.}
\vspace*{-0.2cm}
\end{table}

\subsection{Semantic tag analysis}
\label{sec:semanalysis}

We are also interested in finding out \emph{why} the character-level representations help improve performance. As a start, we investigate on what type of sentences and semantic phenomena the character representations are the most beneficial. We introduce a novel method of analysis: selecting subsets of sentences based on the occurrence of certain semantic tags. In the PMB release, each token is also annotated with a semantic tag, which indicates the semantic properties of the token in the given context \cite{semantic-tagset:17}. This allows us to easily select all sentences that contain certain (semantic) phenomena and evaluate the performance of the different models on those sentences.\footnote{Note that this method of analysis can easily be used for other NLP tasks as well, the only requirement being that a semantic tagger has to be used to get the semantic tags.}

The selected phenomena and corresponding F-scores for our four best models (see Table~\ref{tab:finalres}) are shown in Table~\ref{tab:semtag}.\footnote{See Appendix~\ref{sec:appsemtags} for the list of semtags per category.} Our best model (+ch+sem) has the best performance on six of the seven phenomena selected, even though the differences are small. The character-level representations seem to help across the board; the +char models improve on the baseline (\bert) in almost all instances.

For \emph{Numerals} and \emph{Named Entities} we expected the characters to help specifically, since (i) \bert{} representations might not be as optimal for all individual numerals \citep{wallace-etal-2019-nlp}, and (ii) the character representations might attend more to capital letters, which often indicate the presence of a named entity. Indeed, the character representations clearly help for \emph{Numerals}, but less so for \emph{Named Entities}. Of course, this analysis only scratched the surface as to why the character-level representations improve performance. We leave a more detailed investigation to future work.

\subsection{Sentence length analysis}
\label{sec:lenanalysis}

\begin{figure}[!t]
  \includegraphics[width=\columnwidth]{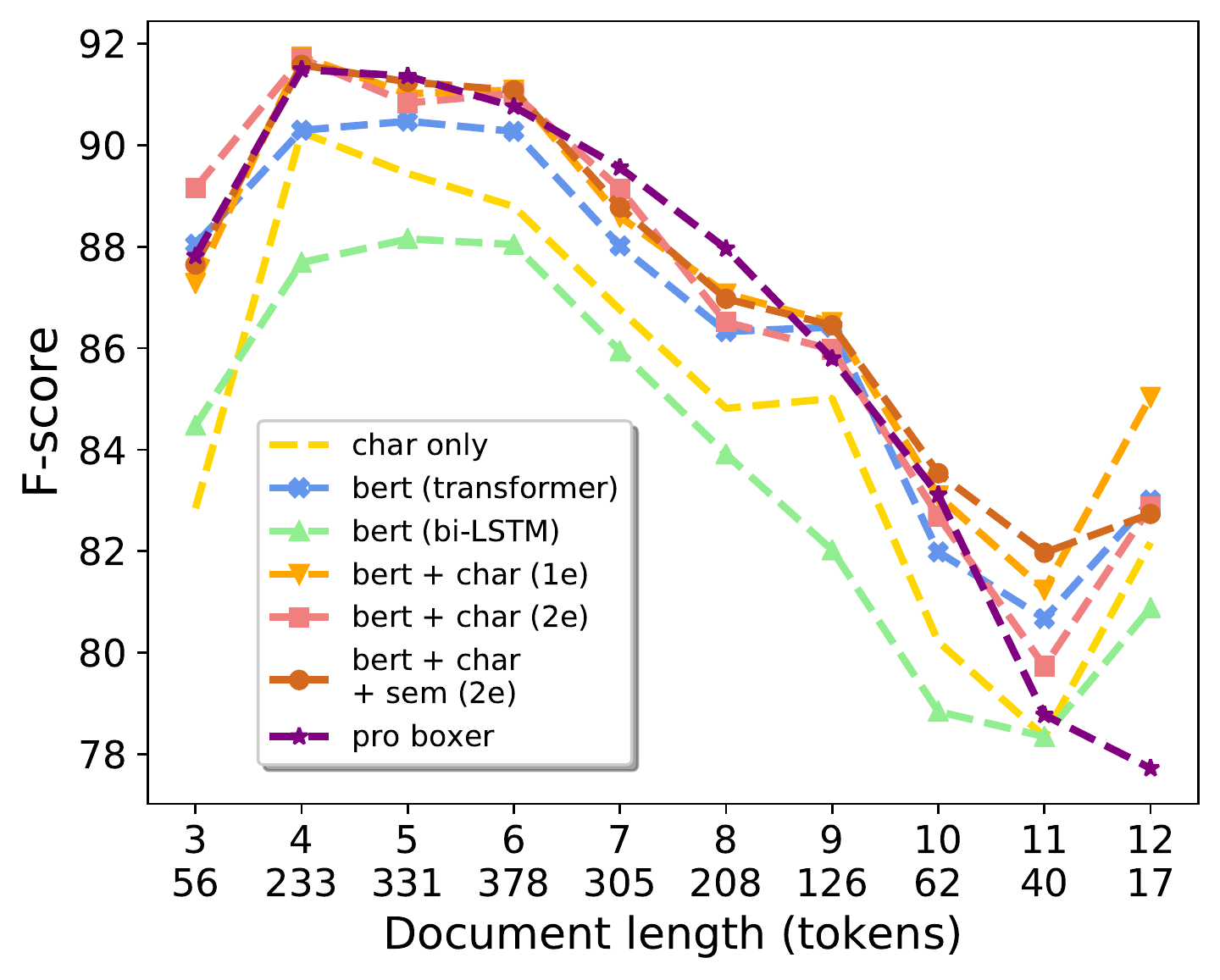}
  \caption{F-scores over document length (tokens) on the combined English dev and test set of 3.0.0. X-axis shows document length (top) and the number of documents for that length (bottom).\label{fig:senlen}}
  \vspace*{-0.2cm}
\end{figure}

We are also interested in finding out which model performs well on longer documents. 
When the Transformer model was introduced, one of the advantages was less decrease in performance for longer sentences \citep{transformer:17}. Also, since Boxer is partly rule-based and not trained in an end-to-end fashion, it might be able to handle longer sentences better. Figure~\ref{fig:senlen} shows the performance over sentence length for seven of our trained systems. We see a similar trend for all models: a decrease in performance for longer sentences.
We also create a regression model that predicts F-score, with as predictors parser and document length in tokens, similar to \citet{drstacl:18}. We do not find a significant interaction of any model with sentence length, i.e., none of the models decreases significantly less or more than any other model.

\begin{figure}[!t]
  \includegraphics[width=\columnwidth]{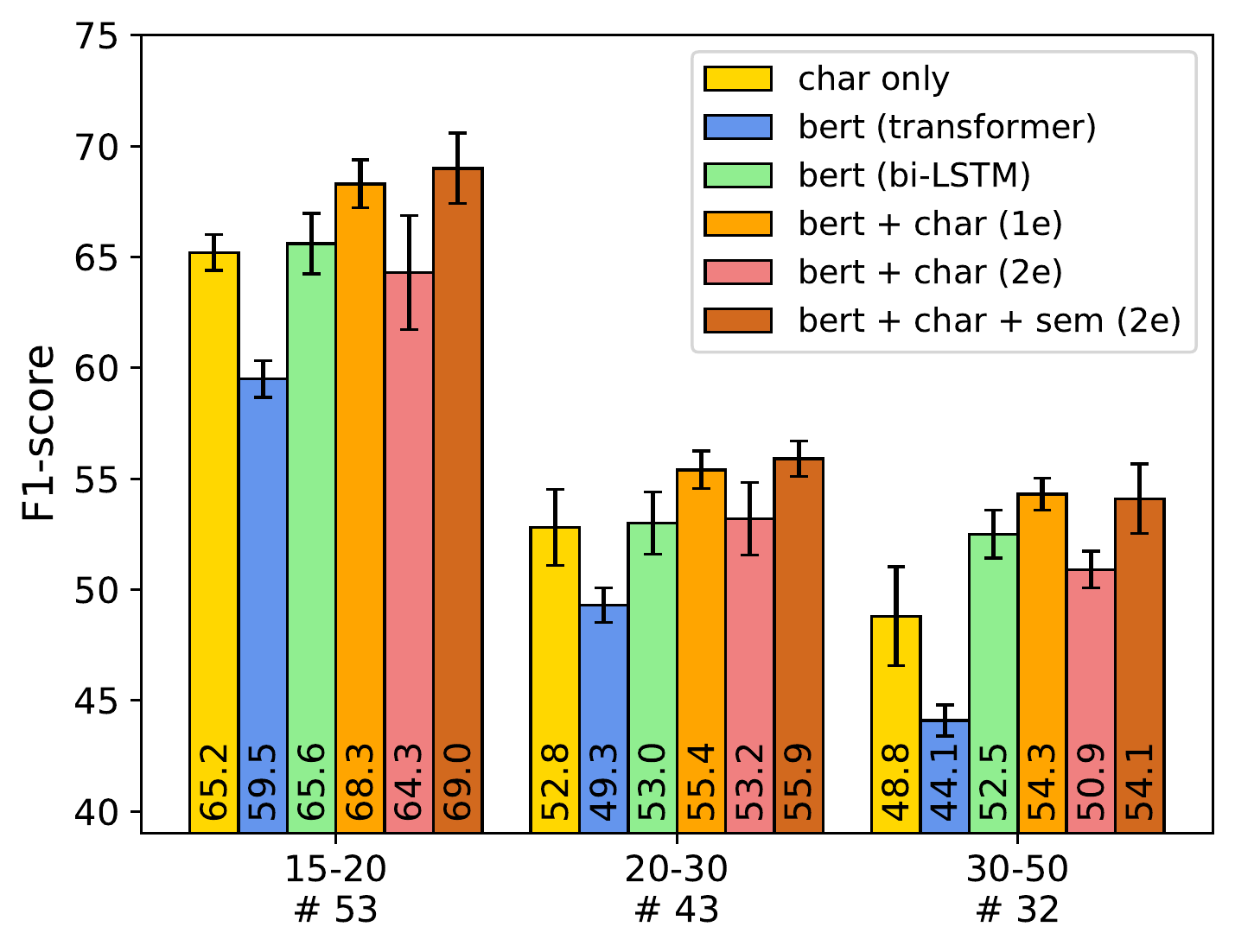}
  \caption{F-scores over document length (tokens) on the silver standard evaluation set of longer documents. X-axis shows the sentence length bins (top) and the number of documents for that length (bottom).\label{fig:silvereval}}
  \vspace*{-0.2cm}
\end{figure}

To get some idea how well our models would do on longer (possibly multi-sentence) documents, we create a new evaluation set. We select all silver documents with 15 or more and less than 51 tokens that have at least the semtagging or CCG layer marked as gold standard. This resulted in a set of 128 DRSs, which should contain the higher quality silver documents. We retrain our models with those sentences removed and plot the performance over sentence length in Figure~\ref{fig:silvereval}. We see that performance still decreases for longer sentences, though not as much after 30 tokens per document. The Transformer model does not seem to catch up with the bi-LSTM models, even for longer documents. The addition of characters is still beneficial for longer documents, though only in one encoder. 

\subsection{Discussion}

We found that adding character-level representations generally improved performance, though we did not find a clear preference for either the one-encoder or two-encoder model. 
We believe that, given the better performance of the two-encoder model on the fairly short documents of the non-English languages (see Figure~\ref{fig:nonen}), this model is likely the most useful in semantic parsing tasks with single sentences, such as SQL parsing \citep{zelle1996learning, iyer-etal-2017-learning, finegan-dollak-etal-2018-improving}, while the one encoder char-CNN model has more potential for tasks with longer sentences/documents, such as AMR \citep{amr:13}, UCCA \citep{ucca:13} and GMB-based DRS parsing \citep{GMB:2017,neural_drs_gmb:18, liu-etal-2019-acl}. The latter model also has more potential to be applicable for other (semantic parsing) systems as it can be applied to all systems that form token-level representations from a document. In this sense, we hope that our findings here are also applicable for other, more structured, encoder-decoder models developed for semantic parsing (e.g., \citealp{yin-neubig-2017-syntactic,krishnamurthy-etal-2017-neural,coarsefine:18,liu-etal-2019-acl}).

An unexpected finding is that the \bert{} models outperformed the larger \roberta{} models. In addition, it was even preferable to use \bert{} only as initial token embedder, instead of fine-tuning using the full model. 
Perhaps this is an indication that certain NLP tasks cannot be solved by simply training ever larger language models.
Moreover, the Transformer model did not improve performance for any of the input representations, while being harder to tune as well. We are a bit hesitant with drawing strong conclusions here, though, since we only experimented with a vanilla Transformer, while recent extensions (e.g., \citealp{universaltransformer:19,guo-etal-2019-star,press-etal-2020-improving}) might be more promising for smaller data sets.

\section{Conclusion}

We performed a range of experiments on Discourse Representation Structure Parsing using neural sequence-to-sequence models, in which we vary the neural representation of the input documents. We show that, not surprisingly, using pretrained contextual language models is better than simply using characters as input (\textbf{RQ1}). However, characters can
still be used to improve performance, in both a single encoder and two encoders (\textbf{RQ2}). The improvements are larger than using individual sources of linguistic information, and performance still improves in combination with these sources (\textbf{RQ3}). The improvements are also robust across different languages models, languages and data sets (\textbf{RQ4}) and improve performance across a range of semantic phenomena (\textbf{RQ5}). These methods should be applicable to other semantic parsing and perhaps other natural language analysis tasks.

\section*{Acknowledgements}

This work was funded by the NWO-VICI grant ``Lost in Translation---Found in Meaning'' (288-89-003). 
The Tesla K40 GPU used in this work was kindly donated to us by the NVIDIA Corporation. 
We thank the Center for Information Technology of the University of Groningen for providing access to the Peregrine high performance computing cluster. 
We also want to thank three anonymous reviewers and our colleagues Lasha Abzianidze, Gosse Minnema, Malvina Nissim, and Chunliu Wang for their comments on earlier versions of this paper.

\bibliography{emnlp2020}
\bibliographystyle{acl_natbib}

\appendix

\section{Char-CNN}
\label{app:charcnn}

\begin{figure}[!htb]
  \includegraphics[width=0.45\textwidth]{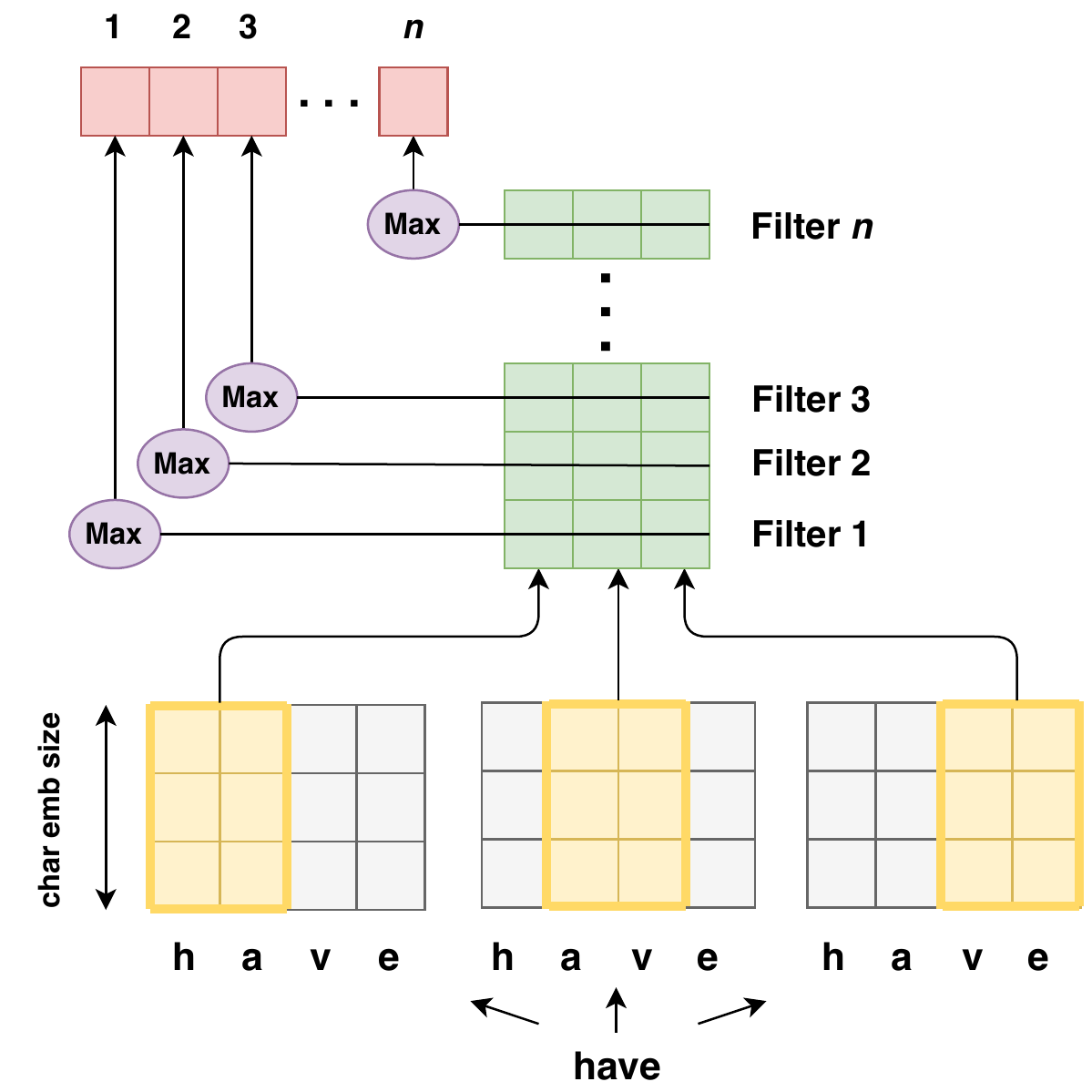}
  \caption{Overview of the char-CNN encoder, encoding the word \emph{have} with bigrams (width = 2) for \emph{n} filters.\label{fig:charcnn}}
\end{figure}

Figure~\ref{fig:charcnn} shows a schematic overview of using the char-CNN \citep{kim2016character} to encode the word \emph{have} with a width of 2.  A width of 2 selects the bigrams \emph{ha}, \emph{av} and \emph{ve}, returning a scalar for each bigram operation, which in turn form a vector $\vect{f}_1$ for filter 1. We then take the max value of this vector to obtain the first value of our width 2 ($w_2$) char-CNN embedding $\vect{e}_{\mathrm{w2}_1}$. The final vector $\vect{e}_{w2}$ is thus of length $n$.

\section{Experimental settings}
\label{sec:hyperappendix}

\inlineheader{Tuning} Table~\ref{tab:hyperspec} gives an overview of the hyperparameters we used and/or experimented with in the tuning stage. This table only gives an overview of the settings for the \bertbase{} model, though the settings for the other representations (described in Section~\ref{sec:lms}) are usually very similar. We performed manual tuning, selecting the settings with the highest F1-score. The number of tuning runs was between 10 and 40 for each representation type and model combination (see Table~\ref{tab:charvsbert}). Output, evaluation (containing F1-scores, standard deviation and confidence interval) and configuration files for our four best models (see Table~\ref{tab:finalres}) are available here: \url{https://github.com/RikVN/Neural_DRS/}.

\inlineheader{Data filtering}We filtered ill-formed DRSs from the PMB data sets, which only occurs for silver and bronze data ($<$ 0.1\% of DRSs). For the bi-LSTM models, the filtering of source and target tokens (see  Table~\ref{tab:hyperspec}) only filters out three very large documents from training. This was done for efficiency and memory purposes, it did not make a difference in terms of F1-score. However, for the Transformer model this improved F1-score by around $0.5$.

\inlineheader{Training time and model size} A single run of the baseline \bert{} model takes about 5 hours to train on a single NVIDIA V100 GPU, with around 17 million trainable parameters. Adding character-level representations in one encoder (using the char-CNN) uses around 55 million trainable parameters, with a runtime of around 6 hours. Using a two encoder setup increases this to around 8 hours, but with only 34 million trainable parameters.

\inlineheader{New evaluation set} When training models that are evaluated on the silver-standard evaluation set of longer documents, we do not perform fine-tuning on the gold standard data. Also, we run Counter with the \texttt{--default-sense} setting (not punishing models that get the word sense wrong), since the word senses of the evaluation set are not gold standard. This has a similar increase of around $1.0$ for all models.

\begin{table}[!h]
\vspace{1.8cm}
\centering
\setlength{\tabcolsep}{4pt}
\resizebox{\columnwidth}{!}{
\begin{tabular}{llll}
\toprule
\textbf{Parameter} & \textbf{LSTM} & \textbf{Transf.} & \textbf{Range} \\
\midrule
Hidden RNN size & 300 & NA & 200 - 600 \\
Decoder RNN size & 300 & NA & 300 \\
\midrule
Num heads & NA & 6 & 2, 4, 6, 10 \\
hidden\_dim & NA & 300 & 300 - 600 \\
ff\_hidden\_dim & NA & 900 & 300 - 1200 \\
dropout: layer        & NA               & 0.1                  & 0.1, 0.2                \\
\hspace{1.38cm} residual              & NA               & 0.2                  & 0.1, 0.2                \\
\hspace{1.38cm}  attention             & NA               & 0.1                  & 0.1, 0.2                \\
\midrule
target emb dim & 300 & 300 & 300 (\glove{}) \\
max src tokens & 125 & 50 & 30 - no max \\
max tgt tokens & 1160 & 560 & 300 - no max \\
layers & 1 & 6 & 1-3 LSTM, 1-10 Trans \\
max\_norm & 3 & 3 & 3, 4, 5 \\
scale\_grad\_by\_freq & False & False & True/False \\
label\_smoothing & 0.0 & 0.1 & 0.0, 0.05, 0.1, 0.2 \\
beam size & 10 & 10 & 10 \\
max decoding steps & 1000 & 500 & 500, 1000 \\
scheduled sampling & 0.2 & 0.0 & 0.0, 0,1, 0.2, 0.3, 0.4 \\
batch size & 48 & 32 & 12, 24, 32, 48, 64, 128 \\
optimizer & adam & adam & adam, sgd, BertAdam \\
learining rate & 0.001 & 0.0002 & 0.0001 - 0.01 \\
grad norm & 0.9 & 0.9 & 0.7 - 0.95 \\
min target occ & 3 & 3 & 1, 3, 5, 10, 20 \\
\bottomrule
\end{tabular}
}
\caption{\label{tab:hyperspec}An overview of the hyperparameters used for the LSTM and Transformer architecture, that use the \bertbase{} representations. Parameters not specified are left at their default value.}
\end{table}

\newpage

\section{Semantic tag selection}
\label{sec:appsemtags}

\begin{table}[!htb]
\centering
\renewcommand{\arraystretch}{1.2}
\setlength{\tabcolsep}{5pt}
\resizebox{\columnwidth}{!}{
\begin{tabular}{ll}
\toprule  
\bf Modality &   \semtag{NOT} \semtag{NEC} \semtag{POS}  \\
\bf Logical  & \semtag{ALT} \semtag{XCL} \semtag{DIS} \semtag{AND} \semtag{IMP} \semtag{BUT} \\ 
\bf Pronouns  & \semtag{PRO}   \semtag{HAS}   \semtag{REF}   \semtag{EMP} \\
\bf Attributes &  \semtag{QUC}   \semtag{QUV}   \semtag{COL}   \semtag{IST}   \semtag{SST}  \\
&  \semtag{PRI}   \semtag{DEG}   \semtag{INT}   \semtag{REL}   \semtag{SCO} \\
\bf Comparatives &  \semtag{EQU}   \semtag{APX}   \semtag{MOR}   \semtag{LES} \\
&  \semtag{TOP}   \semtag{BOT}   \semtag{ORD}  \\
\bf Named entities & \semtag{PER} \semtag{GPE} \semtag{GPO} \semtag{GEO} \semtag{ORG} \semtag{ART} \\
&  \semtag{HAP} \semtag{UOM} \semtag{CTC} \semtag{LIT} \semtag{NTH}  \\
\bf Numerals  & \semtag{QUC} \semtag{MOY} \semtag{SCO} \semtag{ORD} \semtag{DAT} \\
&  \semtag{DOM} \semtag{YOC} \semtag{DEC} \semtag{CLO} \\

\midrule 
\end{tabular}
}
\caption{\label{tab:semtaglist}Semantic tags that were used to select sentences that contain a certain phenomenon. The example sentence in Table~\ref{tab:ling} is included in the categories \emph{Modality}, \emph{Pronouns}, \emph{Named Entities} and \emph{Numerals} .}
\end{table}

\begin{table*}[!t]
\centering
\setlength{\tabcolsep}{4pt}
\resizebox{\textwidth}{!}{
\begin{tabular}{l|rrrr|rrrr|rrrr|rrrr}
\toprule
& \multicolumn{8}{c|}{\textbf{PMB release 2.2.0}} & \multicolumn{8}{c}{\textbf{PMB release 3.0.0}} \\
\midrule
&  \multicolumn{4}{c|}{\textbf{Development set}} & \multicolumn{4}{c}{\textbf{Test set}} & \multicolumn{4}{|c|}{\textbf{Development set}} & \multicolumn{4}{c}{\textbf{Test set}} \\
\midrule
& \textbf{bert} & \textbf{+ch} & \textbf{+ch} & \multicolumn{1}{c|}{\textbf{+ch}} & \textbf{bert} & \textbf{+ch} & \textbf{+ch} & \multicolumn{1}{c|}{\textbf{+ch}} & \textbf{bert} & \textbf{+ch} & \textbf{+ch} & \multicolumn{1}{c|}{\textbf{+ch}} & \textbf{bert} & \textbf{+ch} & \textbf{+ch} & \multicolumn{1}{c}{\textbf{+ch}} \\
& & \bf(1e) & \bf(2e) & \bf +sem & & \bf(1e) & \bf(2e) & \bf +sem & & \bf(1e) & \bf(2e) & \bf +sem & & \bf(1e) & \bf(2e) & \bf +sem \\
\midrule
\bf Prec   & 87.3             & 87.8             & 87.4             & 87.6  & 89.8             & 89.9             & 89.9            & 89.5      & 88.8             & 88.9             & 89.3             & 89.5         & 90.0             & 90.6             & 90.3             & 90.5       \\     
\bf Rec & 83.6             & 84.4             & 83.6             & 83.5  & 86.2             & 86.7             & 86.4            & 86.0   & 86.4             & 87.3             & 86.9             & 87.2              & 87.1             & 87.9             & 87.6             & 88.0       \\      
\bf F1  & 85.4             & 86.1             & 85.5             & 85.5   & 87.9             & 88.3             & 88.1            & 87.7   & 87.6             & 88.1             & 88.1             & 88.4              & 88.5             & 89.2             & 88.9             & 89.3    \\         
\bf Operators & 94.7             & 95.2             & 94.7             & 94.4  & 94.8             & 94.7             & 94.4            & 94.7 &  95.0             & 95.4             & 95.4             & 95.7          & 95.7             & 95.7             & 95.7             & 96.1  \\           
\bf Roles & 88.0             & 88.4             & 88.2             & 88.0  & 90.3             & 90.3             & 90.5            & 89.8 &   89.0             & 89.0             & 89.2             & 89.9          & 89.4             & 90.1             & 89.9             & 90.0  \\           
\bf Concepts & 83.9             & 84.5             & 84.0             & 84.8  & 87.4             & 87.9             & 87.6            & 87.4       & 84.7             & 84.9             & 85.6             & 85.4        & 87.3             & 87.9             & 87.4             & 87.7  \\           
\quad \bf Nouns & 90.8             & 91.5             & 91.1             & 91.4 & 92.4             & 92.8             & 92.4            & 92.5  & 90.6             & 91.0             & 91.4             & 91.5               & 92.0             & 92.5             & 91.8             & 92.5   \\          
\quad \bf Verbs & 65.6             & 65.4             & 64.8             & 67.6  & 75.7             & 76.4             & 76.3            & 75.5  & 69.1             & 68.9             & 70.4             & 69.2              & 75.3             & 76.0             & 76.4             & 75.3   \\          
\quad \bf Adjectives & 70.4             & 74.0             & 72.7             & 71.5  & 70.9             & 72.3             & 70.8            & 71.5  & 76.1             & 75.3             & 76.6             & 75.5          & 75.8             & 77.5             & 76.2             & 76.0   \\          
\quad \bf Adverbs & 90.0             & 67.7             & 83.3             & 63.3 & 70.0             & 71.7             & 73.3            & 61.0  & 78.1             & 77.7             & 78.7             & 80.1              & 88.0             & 88.2             & 87.7             & 88.9   \\          
\quad \bf Events & 66.7             & 67.3             & 66.5             & 68.4  & 74.8             & 75.7             & 75.4            & 74.7 & 70.8             & 70.5             & 71.9             & 70.7             & 75.4             & 76.3             & 76.4             & 75.4  \\     
\midrule
\bf Perfect sense & 87.3             & 88.1             & 87.6             & 87.4  & 89.3             & 89.7             & 89.5            & 89.1   & 89.6             & 90.3             & 90.2             & 90.4        & 91.6             & 92.2             & 92.0             & 92.1  \\  
\bf Infreq. sense & 50.5 & 50.5 & 46.7 & 52.3 & 57.2 & 58.3 & 58.8 & 59.1 & 54.9 & 57.6 & 56.5 & 56.0      &   62.0          &   62.8          &    62.7          & 63.1   \\
\midrule
\bf F1 std dev & 0.30              & 0.30              & 0.17             & 0.05  & 0.22             & 0.22             & 0.16            & 0.19    & 0.19 & 0.25 & 0.30 & 0.34         & 0.26             & 0.24             & 0.29             & 0.22  \\           
\bf F1 confidence & 85.0 & 85.6 & 85.2 & 85.4 & 87.6 & 88.0 & 87.9 & 87.5 & 87.3 & 87.8 & 87.7 & 87.9  & 88.2   & 88.9  & 88.5  & 89.0 \\ 
\bf \quad interval & 85.8 & 86.5 & 85.7 & 85.5 & 88.2 & 88.6 & 88.3 & 88.0 & 87.9 & 88.5 & 88.5 & 88.8               & 88.9     &  89.5  & 89.4  & 89.6 \\
\midrule
\bf \# illformed & 0.4              & 0.0              & 0.2              & 0.2 & 0.2              & 0.0              & 0.2             & 0.0          & 3.2              & 0.8              & 2.8              & 2.0        & 4.6              & 3.0              & 2.8              & 2.0       \\       
\bf \# perfect (avg) & 235.4            & 237.4            & 239.0            & 239.8 & 267.0            & 265.8            & 266.4           & 267.2  & 336.2            & 350.6            & 352.4            & 352.8      & 358.0            & 372.4            & 365.0            & 367.8     \\       
\bf \# perfect (all 5) & 180              & 187              & 183              & 188  & 206              & 213              & 212             & 205 & 212              & 238              & 229              & 226   & 242              & 255              & 239              & 241     \\         
\bf \# zero (avg) & 4.4              & 3.4              & 4.2              & 4.2 & 1.6              & 1.8              & 1.2             & 1.8    & 6.6              & 3.6              & 5.0              & 3.6         & 5.0              & 3.2              & 3.6              & 2.6    \\          
\bf \# zero (all 5) & 4                & 3                & 3                & 3 & 1                & 1                & 0               & 1   & 2                & 2                & 1                & 1          & 0                & 0                & 0                & 0     \\           
\bf \# same (all 5) & 368              & 398              & 379              & 384   &   356              & 368              & 361             & 352    & 347              & 387              & 386              & 365       & 364              & 378              & 361              & 361   \\           
                
\bottomrule
\end{tabular}
}
\caption{Detailed Counter scores for our models on the English dev and test sets of release 2.2.0 and 3.0.0. All scores are averages of 5 runs. Scores are produced by using \drseval{}.\label{tab:detailed}}
\end{table*}

\section{Detailed scores}
\label{app:detailed}

Table~\ref{tab:detailed} shows the detailed F-scores for the English dev and test sets of release 2.2.0 and 3.0.0. \emph{Infreq. sense} is the F-score on all concept clauses that were not the most frequent sense for that word in the training set (e.g., \texttt{be.v.01}, \texttt{like.v.03}). \emph{Perfect sense} is the F-score when we ignore word senses during matching, i.e., \texttt{be.v.01} can match with \texttt{be.v.02}. The last 9 rows are not in the original detailed Counter scores, but are produced by \drseval. Character-level representations help to produce fewer ill-formed and more perfect DRSs, especially on 3.0.0.

\section{Sentence analysis}
\label{app:sents}

Table~\ref{tab:worstsents} shows the sentences for which our best model (on 3.0.0 English dev) produced the lowest quality DRSs, with a possible explanation. In Table~\ref{tab:bestsents}, we show the sentences for which our best model has the best performance (relative to the \bertonly{} baseline model). It is harder to give an explanation in this case, though we indicate which clauses were (in)correctly predicted by the models.

\begin{table*}[!t]
\setlength{\tabcolsep}{5pt}
\resizebox{0.95\textwidth}{!}{
\begin{tabular}{lrl}
\toprule
\textbf{Document} & \textbf{F1} & \textbf{Comment} \\
\midrule
Look out! & 0.00 & Imperative \\
The dove symbolizes peace. & 0.13 & Condition + consequence    \\
HBV Union Criticizes Deutsche Bank & 0.25 & Two multi-word expressions   \\
You can buy stamps at any post office. & 0.32 & Possibility (can) and quantifier (any)   \\
Fire burns. & 0.33  & Generic, short  \\
How's Lanzarote? & 0.36 &  How-question  \\
I'd better drive you home. & 0.37 & Necessity, infrequent sense of drive   \\
What a lot of books! Do they belong to the university library? & 0.38 & Multi-sentence   \\
Maybe he is Italian or Spanish. & 0.40 & Possibility and conjunction    \\
I always get up at 6 o'clock in the morning. & 0.40 & Necessity + clocktime   \\
\bottomrule
\end{tabular}
}
\caption{Sentences of the English 3.0.0 dev set for which our best model (+char +sem) produced the \textbf{worst} DRSs.\label{tab:worstsents}}
\vspace{0.5cm}
\resizebox{\textwidth}{!}{
\begin{tabular}{lrl}
\toprule
\textbf{Document} & \textbf{Diff} & \textbf{Comment} \\
\midrule
Fish surface for air.   &   0.554   & Correctly produced \texttt{Goal}  \\
Oil this bicycle.  &   0.482   & Correctly produced \texttt{oil} as a verb  \\
I'm fed up with this winter, I want spring right now!  &   0.404  & Correctly produced \texttt{CONTINUATION} and \texttt{Pivot}  \\
He's Argentinian.   &  0.386  & \bertonly{} failed to produce \texttt{country} and \texttt{Name}  \\
Alas!   &  0.364  & Odd sentence, but correctly produced \texttt{state.v.01}  \\
Fire burns.   &  0.300  & Bad performance for both, \bertonly{} got a score of 0.0  \\
All journeys begin with a first step.  &   0.300  & \bertonly{} produced a lot of non-matching clauses   \\
How heavy you are!   &  0.299  & \bertonly{} produced a lot of non-matching clauses   \\
One plus two is equal to three.   &  0.252  & Correctly produced \texttt{summation.n.04}  \\
He's not like us.  &   0.246  & Correctly produced \texttt{Theme} and \texttt{Co-Theme}   \\
\bottomrule
\end{tabular}
}
\caption{Sentences of the English 3.0.0 dev set for which our best model (+char +sem) produced the \textbf{best} DRSs, relative to the \bertonly{} baseline.\label{tab:bestsents}}
\end{table*}

\end{document}